\documentclass[10pt,conference]{IEEEtran}
\usepackage{cite}
\usepackage[pdftex]{graphicx}
\usepackage{amsmath}
\usepackage[utf8]{inputenc}
\usepackage[ruled, lined, longend, linesnumbered]{algorithm2e}
\usepackage{array}
\ifCLASSOPTIONcompsoc
  \usepackage[caption=false,font=normalsize,labelfont=sf,textfont=sf]{subfig}
\else
  \usepackage[caption=false,font=footnotesize]{subfig}
\fi
\usepackage{url}
\usepackage{listings}
\usepackage{amsthm}
\usepackage{amssymb}
\usepackage{xspace}
\usepackage{proof}
\usepackage{color}

\providecommand{\norm}[1]{\lVert#1\rVert}

\newcommand*\vect[1]{\bar{#1}}

\def\realnum{\mathbb{R}}
\def\naturalnum{\mathbb{N}}

\def\case{{\bf case}}

\def\nn{f}

\def\inputdist{D}
\def\lips{k}
\def\true{{\bf T}}
\def\false{{\bf F}}

\def\prodnn{pf}
\def\product{{\tt ConstructProduct}}

\def\absint{{\tt AbstractInterpret}}
\def\poly{{ poly}}
\def\err{{ err}}
\def\sample{{\tt SampleAndEstimate}}

\begin{document}
\title{Robustness of Neural Networks:\\A Probabilistic and Practical Approach}
% \title{Practical Robustness Checking for Neural Networks using a
% Probabilistic Approach}


\newcommand{\gt}{\raisebox{4pt}{\includegraphics[height=7pt]{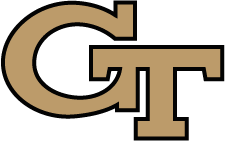}}}
\newcommand{\msr}{\raisebox{5pt}{\includegraphics[height=7pt]{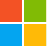}}}

\author{\IEEEauthorblockN{Ravi Mangal\gt \qquad
Aditya V. Nori\msr \qquad
Alessandro Orso\gt}

\IEEEauthorblockA{
	\begin{tabular}{ccc}
		{\small \vspace{-4pt}}\\	
		\gt Georgia Institute of Technology & \hspace{8pt} & \msr Microsoft Research\\
		Atlanta, GA, USA 30332-0765 && Cambridge, CB1 2FB, UK \\
		\{rmangal3, orso\}@gatech.edu && adityan@microsoft.com \\
	\end{tabular}}}

\maketitle

\begin{abstract}
  Neural networks are becoming increasingly prevalent in software, and
  it is therefore important to be able to verify their behavior.
  Because verifying the correctness of neural networks is extremely
  challenging, it is common to focus on the verification of other
  properties of these systems. One important property, in particular,
  is robustness. Most existing definitions of robustness, however,
  focus on the worst-case scenario where the inputs are adversarial.
  Such notions of robustness are too strong,
  and unlikely to be satisfied by---and verifiable
  for---practical neural networks. Observing that real-world inputs
  to neural networks are drawn from non-adversarial probability distributions,
  we propose a novel notion of robustness: probabilistic
  robustness, which requires the neural network
  to be robust with at least $(1-\epsilon)$ probability with respect
  to the input distribution. This
  probabilistic approach is practical and provides a principled way of
  estimating the robustness of a neural network. We also present an
  algorithm, based on abstract interpretation and importance sampling,
  for checking whether a neural network is probabilistically
  robust. Our algorithm uses abstract interpretation to approximate
  the behavior of a neural network and compute an overapproximation of
  the input regions that violate robustness. It then uses importance
  sampling to counter the effect of such overapproximation and compute
  an accurate estimate of the probability that the neural network
  violates the robustness property.
\end{abstract}

\begin{IEEEkeywords}
neural networks; probabilistic; robustness
\end{IEEEkeywords}

%!TEX root = paper.tex

\section{Introduction}
\label{sec:introduction}

Neural networks are increasingly becoming an important computational
component of modern software. With their widespread adoption, it has
become essential that we ensure (or at least gain confidence in) the
correctness of neural networks, as we do with traditional programs.
However, providing formal specifications of correctness is an even
harder task for neural networks than for traditional programs, as neural
networks are explicitly designed for the purpose of learning patterns
in training data that are not easily apparent to humans.

Despite the difficulty of specifying the concept of correctness for
neural networks, there are some important properties that such
networks should satisfy. In particular, in recent years, researchers
have observed certain undesirable neural network behaviors, including
susceptibility to input perturbations \cite{Szegedy13},
unfairness of neural network outcomes \cite{Dwork12,Albarghouthi17},
and leakage of private information (confidentiality and integrity
issues) \cite{Shokri15,Abadi16}.  In this work, we focus on the property
of robustness of neural networks to input perturbations.

An important concept, in the context we target, is \textit{input perturbation}: 
a subtle perturbation of an input such that the
behavior of the neural network is correct on the unperturbed input but
incorrect on the perturbed one. 
Existing literature has focused on the worst-case scenario where
the perturbations are adversarial, without regard to whether
such adversarial inputs are likely to be generated by real-world processes.
Accordingly, a variety of adversarial perturbations/attacks, and defenses against such attacks, have been
proposed (see \cite{Yuan17} for a survey). Further, a variety of
formal definitions of adversarial robustness have been
presented. Broadly speaking, these
adversarial formulations can be classified into two main groups: (i) \textit{local
robustness} and (ii) \textit{global robustness}.

Intuitively, \textit{local robustness} \cite{Bastani16,Huang17,Katz17}
is defined for a given input $x$ and states that the neural network
should produce the same result (e.g., label) for $x$ and for
all inputs $x'$ within a ball of radius $\delta$ centered at
$x$. (Notice that this definition relies on a suitable distance metric
defined over the input space.) 
The requirement to be robust for all inputs that are $\delta$-close 
to $x$ (i.e., within distance $\delta$ from $x$)
might make sense under certain threat models. In practice, however, for neural networks
operating in non-malicious settings, this can be too strong a requirement; 
all $\delta$-close inputs may not be equally likely, and violating robustness
for a highly unlikely $x'$ may be considered practically acceptable.
At the same time, since local robustness is only defined for specific inputs and
provides no guarantees for the inputs that have not been considered, it can also be too weak and inherently limited.
%However, its relatively cheaper computational cost, compared
%to global robustness, has made this the property of choice for
%verifying adversarial robustness.
%
\textit{Global robustness}~\cite{Katz17} addresses this issue by further demanding that the
local robustness property be satisfied by all the inputs in the input
space. In addition to being computationally intractable to check, global
robustness is again too strong to be of practical use. 
%Essentially, it
%potentially requires the network to have the same output on every
%possible input.

To address the practical and conceptual limitations of these existing definitions of adversarial robustness for neural networks, we propose a new robustness property, probabilistic
robustness, that is targeted towards non-adversarial settings and
is globally defined.
Our formulation is motivated by two observations: (i) inputs to a
neural network are generated according to an (either known or unknown)\footnote{Although proving probabilistic robustness requires the underlying input distribution to be known, in the absence
of such information we can rely on a standard known distribution.}
underlying real-world probability distribution over the input space; and 
(ii) in non-adversarial settings, we are only interested in
robustness of a neural network for pairs of $\delta$-close inputs that are
likely to be generated in the real-world.
Consequently, instead of proving robustness for either arbitrary
or all $\delta$-close inputs, we propose to prove robustness for pairs of 
$\delta$-close inputs such that their cumulative probability is at least
$(1-\epsilon)$. Such a proof guarantees that, for a random pair of
$\delta$-close inputs drawn from the input distribution, 
the probability that the neural networks violates robustness
is $\epsilon$ at the most. We believe that, compared to local and global
robustness, probabilistic robustness represents a more practical
and efficiently checkable property.
Moreover, unlike local robustness, this property is globally defined.
Finally, the parameter $\epsilon$ is a tunable knob that 
can be used to control the trade-off between computational efficiency
and strength of the property. 
%
% \RM{The above para is not technically accurate. $\epsilon$ is the
%   probability of input pairs, and the underlying distribution is
%   obtained by the product of the input distribution with itself}

The description we just provided gives an intuitive idea of
probabilistic robustness. More formally, \textit{probabilistic
  robustness} can be expressed by the following formula: 
\vspace{-6pt}
\begin{equation*}
	\underset{x,x' \sim D}{\Pr}(\norm{f(x') - f(x)} \leq k * \norm{x' -
	x} \; \big| \; \norm{x' - x} \leq \delta) \geq 1 - \epsilon
\end{equation*}
Here, $f$ stands for the mathematical function represented by the
neural network, and $f(x)$ represents the output generated by the
neural network on input $x$. $\norm{\cdot}$ represents the norm or
distance metric used over the input and output space (assuming that
the same metric is used for both). This definition states that for a
randomly sampled pair of inputs, conditioned on the inputs being
$\delta$-close, function $f$ satisfies the \textit{Lipschitz
  property}; that is, the distance between the outputs is bounded by a
$k$-multiple of the distance between the inputs, with a high probability
($1-\epsilon$). In other words, and more intuitively, probabilistic robustness 
requires that pairs of inputs that are (1) drawn from the real-world and 
(2) close to each other, result in outputs that are similarly close with a high probability.
Note that this definition
does not apply when the output is discrete or categorical, and we
assume that the output is a real vector. This assumption, however,
does not affect the applicability of our technique; as we further explain in
Section~\ref{sec:property}, even neural networks that produce categorical outputs can be treated as producing
a distribution over the class labels as an output (using a soft-max as
a final layer, for instance).

To check whether a given neural network satisfies probabilistic robustness,
a naive verification algorithm would run the network against every $\delta$-close 
pair of inputs and check and record whether the Lipschitz property is
satisfied. It would then compute the total probability of
all the recorded pairs of inputs and check whether this probability is
greater than $(1-\epsilon)$. Obviously, this algorithm would be impractical,
as the input space of neural networks can be arbitrarily large.
%Moreover, eveni computing the total probability of all the input pairs
%satisfying the property is non-trivial.

To make our approach feasible, we present an algorithm that combines
abstract interpretation~\cite{Cousot77} (from programming languages
theory) and importance sampling~\cite{Robert10} (from statistical
theory) and makes the verification of probabilistic robustness
computationally tractable. Abstract interpretation can take as input
(the precise description of) a program with possibly infinite
behaviors and generate a finite, sound, precise, and computable
approximation of the program behaviors; we use abstract interpretation
to approximate the behavior of a neural network without running it on
all possible inputs. Importance sampling is a sampling technique that
helps improve the precision of statistical estimates, while reducing
the number of samples necessary to compute the estimate; we use
importance sampling to estimate the probability of all the input pairs
that satisfy the property.

The contributions of this work are twofold. First, we propose probabilistic robustness, a new non-adversarial  robustness property of neural networks. Second, we present a practical algorithm for checking whether a network satisfies this property. 

% In section \ref{sec:background}, we present some background on neural
% networks, abstract interpretation, and importance sampling, in section
% \ref{sec:property}, we present the probabilistic robustness property
% in detail, we present our algorithm in section \ref{sec:algorithm},
% and finally conclude in section \ref{sec:conclusion}.

%%% Local Variables:
%%% mode: latex
%%% TeX-master: "paper"
%%% End:

%!TEX root = paper.tex

\section{Background}
\label{sec:background}

\begin{figure}
\begin{small}
\[
\begin{array}{rcl}
	f(\vect{x}) &:=&	W \cdot \vect{x} + \vect{b} \\
		& \mid & \case \; E_1:f_1(\vect{x}),...,\case \; E_k:f_k(\vect{x}))\\
		& \mid & f(f'(x)) \\
	E	&:=& E \wedge E \;|\; x_i \geq x_j \;|\; x_i \geq 0 \;|\; x_i < 0 \\

  \end{array}
\]
\vspace{-12pt}
\caption{Definition of CAT functions.}
\label{fig:cat}
\vspace{-12pt}
\end{small}
\end{figure}

\textbf{Neural networks} are functions that map real-valued vector
inputs to real-valued vector outputs. In this paper, we use the
conditional affine transformations (CAT) representation of neural
networks \cite{Gehr18,Bastani16}, shown in Figure~\ref{fig:cat}. CAT
functions consist of functions $f:\realnum_m \mapsto \realnum_n$,
where $m,n \in \naturalnum$, and are recursively defined. Any affine
transformation $f(\vect{x}) \; := \; W \cdot \vect{x} + \vect{b}$, for
matrix $W$ and vector $\vect{b}$, is a CAT function. Conditional
expressions with multiple cases, and composition of CAT functions, are
CAT functions.  There is a straightforward translation of neural
networks with ReLU activation functions~\cite{glorot11} and standard
layer types (e.g., fully connected layer, convolutional layer, and max
pooling layer) into CAT functions (see Gehr et al.'s work
\cite{Gehr18} for details).

\textbf{Abstract interpretation}~\cite{Cousot77} is a
framework for understanding and proving properties about programs with
potentially infinite behaviors. Abstract interpretation
techniques can soundly approximate these behaviors in a finite,
computable way.  Although the details of abstract interpretation are
beyond the scope of this paper, we provide an example to intuitively
explain the approach.  Consider a function
$f:\realnum_m \mapsto \realnum_n$, where $m,n \in \naturalnum$, and a
set $C \subseteq \realnum_n$.  Suppose that we want to find the
largest set $X \subseteq \realnum_m$ such that
$\forall x \in X. f(x) \in C$. If $f$ is invertible, one way to
compute $X$ is by computing $F^{-1}(C)$, where
$F^{-1}:P(\realnum_n) \mapsto P(\realnum_m)$, and
$F^{-1}(Y) = \{x | f(x) \in Y\}$.  $F^{-1}$ is just a lifting of
$f^{-1}$ to be over a set of outputs $Y$, rather than a single output
$y$, and is called the \textit{concrete backward transformer}. If
$F^{-1}$ has an efficient representation, $F^{-1}(C)$ can be computed
efficiently, but $F^{-1}$ itself can be very inefficient (even
non-terminating). Using abstract interpretation, we can design an
\textit{abstract backward transformer} $\hat{F}^{-1}$ such that
computing $\hat{F}^{-1}(\hat{C}) = \hat{X}$ is guaranteed to be
efficient, and $C \subseteq \hat{C}$ and $X \subseteq \hat{X}$ (i.e.,
$\hat{C}$ and $\hat{X}$) are sound overapproximations of $C$ and $X$,
respectively.

\textbf{Importance sampling}~\cite{Robert10} is a sampling technique
for estimating unlikely properties of distributions.  In particular,
if the region in which the property holds has a low probability,
vanilla Monte Carlo sampling is very unlikely to produce points from
within that region; one is forced to either generate a large number of
samples, or accept a very imprecise (large variance) estimate of the
property under consideration. Importance sampling can help in such a
situation.  Instead of sampling from the original distribution, we
(1) sample from a distribution that attaches a high probability to the
region of interest, (2) estimate the property for this new
distribution, and (3) weight this estimate so as to generate the
estimate for the original distribution. In many cases, importance
sampling can help generate precise estimates with much fewer samples
compared to vanilla Monte Carlo sampling.

%%% Local Variables:
%%% mode: latex
%%% TeX-master: "paper"
%%% End:

%!TEX root = paper.tex

\section{Probabilistic Robustness}
\label{sec:property}

As we stated earlier, existing formulations of neural
network robustness are focused on the worst-case (i.e., the adversarial setting).
Practically, these formulations are not only too strong, but also computationally expensive to
verify. Our formulation of probabilistic robustness
aims to find a practical notion of robustness that is suitable for
non-adversarial settings and is computationally efficient to verify.

To contrast our formulation with the existing ones, we first provide a
formal definition of local and global robustness. A neural network
satisfies local robustness at input $x_0$ if the following formula
holds true:
\vspace{-6pt}
\begin{equation*}
	\forall x. \norm{x_0 - x} \leq \delta \implies f(x_0) = f(x)
	\vspace{-4pt}
\end{equation*}
In the formula, $f$ is the mathematical function represented by the
neural network, and $\norm{\cdot}$ is a distance metric defined on the
input space. Intuitively, the formula states that for all inputs in
the ball of radius $\delta$ centered at $x_0$, the network produces
the same output. Note that input $x_0$ must be explicitly provided.
Because there is no principled guidance on which inputs to select,
such inputs are typically selected in an ad-hoc fashion.

Global robustness basically consists of enforcing local robustness for
every input in the input space and can be expressed as follows:
\vspace{-6pt}
\begin{equation*}
	\forall x,x'. \norm{x - x'} \leq \delta \implies f(x) = f(x')
\vspace{-4pt}
\end{equation*}
Because this formula is universally quantified over both $x$ and $x'$,
this property tends to be too strong to be of practical use---most
real-world neural networks are likely to violate it.

In contrast, a neural network satisfies probabilistic robustness if
the following formula holds true,
\vspace{-6pt}
\begin{equation*}
	\underset{x,x' \sim D}{\Pr}(\norm{f(x') - f(x)} \leq k * \norm{x' - x}) \; \big| \; \norm{x' - x} \leq \delta)\geq 1 - \epsilon
\vspace{-4pt}
\end{equation*}
In the formula, $D$ indicates the input distribution. Probabilistic
robustness differs from local and global robustness in two major ways.
\textit{First}, instead of requiring that the neural network produces
equal output on multiple different inputs, this property bounds the
distance between every pair of outputs in terms of the distance
between the corresponding pair of inputs. (A function satisfying this
property over its entire domain is referred to as a Lipschitz
continuous function). \textit{Second}, the property is not established
for arbitrary or all $\delta$-close inputs. Instead, to prove
the property, one needs to establish it for pairs of $\delta$-close inputs with a total
probability of at least $(1-\epsilon)$ with respect to the
distribution $D$. In case the exact underlying distribution is
unknown, which is likely to be the common case, one can prove this
property for some standard
distribution and still infer useful information about the neural
network.  Note that, because the notion of Lipschitz continuity does
not apply to functions with a discrete or categorical output, we
require that the output of the neural network be continuous. However, 
this does not practically restrict the class of neural networks that
we can consider; even neural networks that act as classifiers
typically produce a real-valued vector as output, where each element
$k$ of the vector represents the probability of the input having label
$k$.

%%% Local Variables:
%%% mode: latex
%%% TeX-master: "paper"
%%% End:

%!TEX root = paper.tex

\section{Algorithm}
\label{sec:algorithm}

\vspace{-12pt}
\begin{algorithm}
\caption{Checking Probabilistic Robustness.}
\label{alg:probalg}
\KwIn{$\nn$: Neural network as a CAT function.\\
\quad \quad \quad $\inputdist$: Input distribution.\\
\quad \quad \quad $\epsilon$: Probabilistic error bound.\\
	\quad \quad \quad $\lips$: Lipschitz constant.}
	\KwOut{$\{\true,\false\}$}
%\Begin{
	$\prodnn$ := $\product(\nn)$\;
%	$\phi$ := $\property(\prodnn)$\;
	$\phi$ := $\neg(\norm{f(x') - f(x)} \leq \lips * \norm{x' - x}))$\;
	$\poly$ := $\absint(\prodnn,\phi)$\;
	$\err$ := $0$\;
	\ForEach{$p \in \poly$}{
		$e$ := $\sample(p,\prodnn,\phi,\inputdist)$\;
		$\err$ := $\err + e$\;
	}
	\uIf{$\err > \epsilon$}{
		\Return $\false$\;
	}\uElse{
		\Return $\true$\;
	}
%}
\end{algorithm}
\vspace{-12pt}

Algorithm \ref{alg:probalg} describes the procedure for checking the
probabilistic robustness of a neural network $\nn$. $\nn$ is input to
the algorithm and is expressed in the form of a CAT function (see
Section~\ref{sec:background}). The other inputs to the algorithm are
the probabilistic error bound $\epsilon$, the Lipschitz constant
$\lips$, and the input distribution $\inputdist$. $\inputdist$ can
either be represented as a closed form function, or as a probabilistic
program, depending on the algorithm implementation.  The algorithm
outputs $\true$ (true) if $\nn$ satisfies probabilistic robustness,
and $\false$ (false) otherwise.

Our algorithm frames the problem of checking the probabilistic
robustness of a neural network as a relational program verification
problem \cite{Barthe11}.  Relational verification is defined as
checking program properties or specifications that are expressed over
pairs of program traces.  For instance, probabilistic robustness
requires comparing the outputs ($\norm{f(x') - f(x)}$) generated by a
neural network for pairs of inputs ($\norm{x' - x}$). Such two-trace
properties are also called \textit{hyperproperties} \cite{Clarkson10}.

A majority of program verification and analysis techniques are only
applicable to single-trace properties. To be able to use such
techniques for checking hyperproperties, a standard trick used in
program verification is to construct a product program
\cite{Barthe04}. For a program $P$, a product program is
constructed by creating a copy $P'$ of $P$, where all the variables
are renamed, and composing $P$ and $P'$ together to get program
$P;P'$. A hyperproperty of the original program then corresponds to a
single-trace property of the product program.

The first step of our algorithm is to construct a ``product" neural
network $\prodnn$ (line 1) by encoding two copies of the original
network $\nn$ side by side. Assume that the input and the output of
the original neural network $\nn$ are notated as $\bar{x}$ and
$\bar{y}$, respectively. Then, intuitively, the product neural network
(1) accepts the input $(\bar{x}, \bar{x}')$, (2) independently
processes $\bar{x}$ and $\bar{x}'$, and (3) produces the output
$(\bar{y}, \bar{y}')$, such that $\bar{y}$$=$$\nn(\bar{x})$ and
$\bar{y}'$$=$$\nn(\bar{x}')$. This product construction enables us to use
standard abstract interpretation techniques for checking a
hyperproperty such as robustness. Note that, as we just discussed, any
input for the product neural network represents a pair of inputs for
the original neural network. In the rest of this section, we therefore
use the term input to refer to a product neural network input.

In line~2, the algorithm assigns the temporary name $\phi$ to the
property to be checked, that is, the negation of the Lipschitz
property. The backwards abstract interpreter $\absint$ produces the
set $\poly$ (line 3) as an overapproximation of the set of inputs that
satisfy $\phi$. Since $\phi$ is the negation of the Lipschitz
property, all the inputs NOT in $\poly$ satisfy the Lipschitz
property. Because $\absint$ is based on the powerset polyhedra
abstract domain \cite{Cousot78,Sankaranarayanan06}, which uses a set
of polyhedra to approximate a set of real-valued vectors, the set
$\poly$ produced by the abstract interpreter is a set of input
polyhedra.
%Next, for each input polyhedron $p$ in $\poly$, the algorithm computes
%the probability $v$ of a random input being included in $p$. This
%probability is computed with respect to the input distribution
%$\inputdist$ (line 6). Doing so is equivalent to computing the volume
%of $p$ weighted by the probability distribution $\inputdist$.
%Computing the volume of a polyhedra, weighted by a probability
%distribution, is a well-studied problem \cite{Vempala05,Deloera13},
%and the $\vol$ procedure on line~6 can use any of the existing
%algorithms to compute such volume.

Next, for each input polyhedron $p$ in $\poly$, the algorithm applies importance sampling to
improve the precision of the results. As we discussed above, each
polyhedron $p$ computed through abstract interpretation is an
overapproximation of the set of inputs that satisfy $\phi$ (i.e., the
set of inputs that violate the Lipschitz property). 
%To reduce imprecision, the algorithm samples inputs from within $p$ and uses
%these samples to compute the ratio $e$ of inputs in $p$ that satisfy
%$\phi$. For instance, if the algorithm samples 100 inputs form $p$,
%and 80 of these inputs satisfy $\phi$, the algorithm would compute $e$
%to be 0.8. Then, to get the probability of inputs within $p$
%satisfying $\phi$ the algorithm weighs $e$ by the probability $v$ of
%the polyhedron $p$ and updates $\err$, the total probability of
%satisfying $\phi$ (line 8). 
To reduce imprecision, the algorithm samples inputs from within $p$,
and uses these samples to estimate the probability $e$ of inputs in $p$
satisfying $\phi$. For each sample, the sampling procedure first checks
if the distance between the two elements comprising the sample input 
is more than $\delta$. If so, the sample is rejected. Otherwise, the sample is accepted.
For each
accepted sample, the sampling procedure checks if the sample satisfies $\phi$.
The probability estimate $e$ is the average weighted probability
of the samples satisfying $\phi$, where the weighted probability depends
on the size of $p$ and on the input distribution $\inputdist$.
Finally, after processing all polyhedra,
the algorithm checks the value of $\err$, which is the total probability of
satisfying $\phi$. If $\err$ is greater than
$\epsilon$, the probability of violating the Lipschitz property is
greater than $\epsilon$, neural network $\nn$ is not probabilistically
robust, and the algorithm returns $\false$ (lines 9--10). Otherwise,
$\nn$ satisfies the property, and the
algorithm returns $\true$ (lines 11--12).

%%% Local Variables:
%%% mode: latex
%%% TeX-master: "paper"
%%% End:

%\input{related}
%!TEX root = paper.tex

\section{Conclusion}
\label{sec:conclusion}

We presented probabilistic robustness, a novel formulation of
robustness of neural networks that is practical, yet principled.
Probabilistic robustness guarantees
that a neural network is robust with at least $(1-\epsilon)$
probability, given a real-world input probability distribution.
In contrast to existing notions of robustness, probabilistic robustness
focuses on a non-adversarial setting.
We also presented an algorithm based on abstract interpretation and importance
sampling for checking whether a neural network is probabilistically
robust. We are currently implementing our algorithm and plan to
evaluate the usefulness of our approach on real-world neural networks.

%%% Local Variables:
%%% mode: latex
%%% TeX-master: "paper"
%%% End:

\section{Acknowledgments}
{\small This work was partially supported by 
NSF, under grants CCF-1161821 and 1563991,
DARPA, under contracts FA8650-15-C-7556 and FA8650-16-C-7620,
ONR, under contract N00014-17-1-2895,
and gifts from Google, IBM Research, and Microsoft Research.
We thank the anonymous reviewers for their helpful feedback.}

\bibliographystyle{IEEEtran}
\bibliography{references}

\end{document}